\documentclass{article}
\usepackage{graphicx}
\usepackage{subcaption} 
\usepackage{mwe}
\usepackage{comment}
\usepackage{caption}
\usepackage{appendix}
\usepackage{wrapfig}
\usepackage{authblk}

\usepackage[final]{neurips_2019}

\usepackage[utf8]{inputenc} % allow utf-8 input
\usepackage[T1]{fontenc}    % use 8-bit T1 fonts
\usepackage{hyperref}       % hyperlinks
\usepackage{url}            % simple URL typesetting
\usepackage{booktabs}       % professional-quality tables
\usepackage{amsfonts}       % blackboard math symbols
\usepackage{nicefrac}       % compact symbols for 1/2, etc.
\usepackage{microtype}      % microtypography
\usepackage{graphicx}

\title{SAVEHR: Self Attention Vector Representations for EHR based Personalized Chronic Disease Onset Prediction and Interpretability}

\makeatletter
\renewcommand*{\@fnsymbol}[1]{\ensuremath{\ifcase#1\or 1\or 2\or \ddagger\or
    \mathsection\or \mathparagraph\or \|\or **\or \dagger\dagger
    \or \ddagger\ddagger \else\@ctrerr\fi}}
\makeatother

\author{%
Sunil Mallya$^*$ \thanks{Amazon Web Services Inc, Seattle, WA, USA. \texttt{$^*$smallya@amazon.com}}%
\and
Marc Overhage \thanks{Cerner Corporation, Kansas City, MO, USA.}%
\and
Sravan Bodapati$^1$
\and Navneet Srivastava$^1$ \and Sahika Genc$^1$
}

\begin{document}
\maketitle

\begin{abstract}
Chronic disease progression is emerging as an important area of investment for healthcare providers. As the quantity and richness of available clinical data continue to increase along with advances in machine learning, there is great potential to advance our approaches to caring for patient. An ideal approach to this problem should generate good performance on at least three axes namely, a) perform across many clinical conditions without requiring deep clinical expertise or extensive data scientist effort, b) generalization across populations, and c) be explainable (model interpretability). We present SAVEHR, a self-attention based architecture on heterogeneous structured EHR data that achieves $>$ 0.51 AUC-PR and $>$ 0.87 AUC-ROC gains on predicting the onset of four clinical conditions (CHF, Kidney Failure, Diabetes and COPD) 15-months in advance, and transfers with high performance onto a new population. We demonstrate that SAVEHR model performs superior to ten baselines on all three axes stated formerly.
\end{abstract}

\section{Introduction}

Clinicians record structured data such as diagnosis codes, vitals from lab tests and unstructured data such as clinical notes in electronic health records (EHR) system. Accurately predicting the progression of diseases using aforementioned data from EHR could allow clinicians and patients to make more informed choices, reduce costs, and decrease mortality and morbidity. But challenges in EHR data include, heterogeneity, temporal dependencies, sparseness and incompleteness while being high dimensional. [\cite{jensen2012mining}, \cite{weiskopf2013defining}, \cite{tran2014framework}]. Data-driven approaches for feature selection from EHR have been proposed to address these challenges. [\cite{huang2014toward}, \cite{lyalina2013identifying}, \cite{wang2014unsupervised}]. An initial step in modeling a disease trajectory is to predict its onset. A variety of deep learning approaches to predicting disease onset have been explored including predictions of congestive heart failure \cite{mallya2019effectiveness}, Kidney Failure \cite{perotte2015risk}, Dementia \cite{de2018unsupervised} and Delirium \cite{wong2018development}. High performance while a necessity, validation on a new population and interpretability are key aspects for adoption in a healthcare system. We answer these aspects by proposing SAVEHR which uses self-attention \cite{lin2017structured} on structured EHR data to learn pairwise comorbidities to effectively predict disease onset, and generate personalized feature importance visualizations.

\textbf{Related Work}:
In recent years, Attention mechanisms have made substantial gains in conjunction with RNNs. Attention allows the network to focus on certain regions of data, while perceiving other regions with “low resolution”. As a consequence of that, it facilitates the interpretation of learned representations. We now see attention being applied to healthcare data (clinical notes and structured EHR) as well, To represent the behavior of physicians during an encounter a two-level neural attention model is used by \cite{choi2016retain} focusing on reverse time order of events. EHR events as a temporal matrix by \cite{cheng2016risk} and use CNN based architecture to predict onset of CHF and COPD, to obtain feature importance they aggregate weights of the neurons. To predict outcomes on ICU events, attention is used by \cite{kaji2019attention} and attention is used on clinical notes to detect adverse medical events by \cite{chu2018using}.  Multi-stage attention is used by \cite{Patient2Vec}, where self-attention is applied within a sub-sequence of homogeneous features like medical codes followed by creation of aggregated deep representation to predict outcomes and generate personalized heatmap for a patient that can explain model predictions.

\section{Cohort}
We create one development and one external test cohort for each of the four chronic diseases (CHF, Kidney Failure, Diabetes Type II and COPD) from de-identified, anonymized, structured EHR data that are from two distinct patient populations referred to as P1 (the development cohort) and P2 (the test cohort). We use a 12-month observation window that's between 27 and 15-months from the index date and a prediction window of 15-months. Since, the observation window was fixed across a relatively long time window (12-month), we aggregate frequency counts of diagnosis codes assigned for encounters across a time window (3-month time slices or quarters) similar to that in \cite{Choi2016UsingRN} to facilitate temporal learning. The case-control design, index date selection along with time windows is explained in the Appendix \ref{index_date}. We represent each patient's data with static demographic features (gender, race, age) and sequence of  diagnoses and procedures codes, termed as medical concepts. Feature sequence for a patient $i$ is denoted as $v_q^{(i)}$, where $q$ denotes the quarter of interest, so the higher the subscript value, the closer the quarter is to the index date. The number of case and control patients for each disease is presented in Appendix \ref{tab:popstats}.

%For example $v_1^0$ denotes vector representation of first patient in first quarter, farthest from the index date. This sequence has a homogeneous component $h_i^t$ representing the diagnosis codes assigned to patients during their visits in timeslice $t$. $h_i^t \in R^D$ is a sub-sequence of codes assigned in a given quarter from a universal set of diagnosis codes $M$ such that $h_i^t$ = \{$m_{t_1}$,$m_{t_2}$,$m_{t_3}$,...$m_{t_k}$\}, where $m_{t_i} \in R^D$ is the $i^{th}$ medical code applicable to the patient for a given quarter $t_j$, and there are $k$ codes in the sub-sequence. 

%%%%
\section{SAVEHR Model Architecture}
\begin{wrapfigure}{L}{0.5\textwidth}
\begin{center}
  %\begin{subfigure}[b]{0.55\textwidth}
    \includegraphics[width=0.5\textwidth]{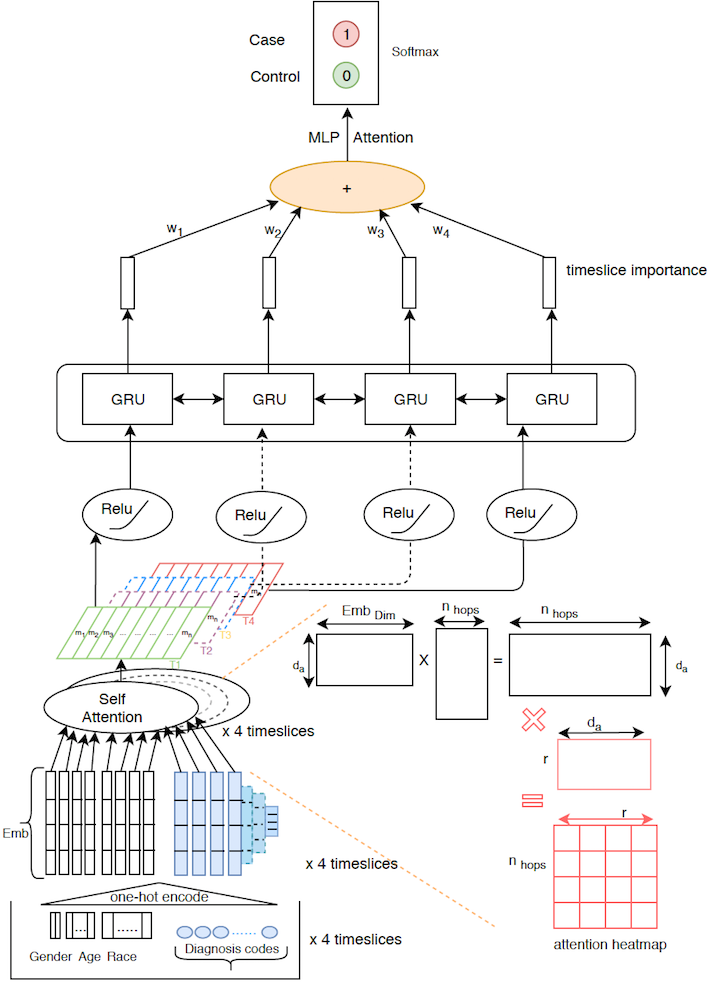}
    \caption{SAVEHR Deep Neural Network }
    \label{fig:architecture}
\end{center}
\end{wrapfigure}
%Disease progression is influenced by comorbidities and we believe that explicitly modeling the co-occurrences of conditions can improve performance.
In this section, we present details and the architecture(Figure~\ref{fig:architecture}) of the proposed self-attention based SAVEHR neural network.  There are three main components of this architecture: 1) self-attention layer for heterogeneous features, followed by 2) a Bi-GRU layer and 3) an MLP attention mechanism. In the following, we describe each component and their contribution to the classification task in detail.

\textbf{Self-Attention with Heterogeneous features:} Self-Attention relates elements at different positions from a single sequence by computing the attention between each pair of inputs, $x_i$ and $x_j$. Non-categorical information such as age is converted into categorical feature by binning, while race $R_i$ and gender $G_i$ are integer encoded. The final feature representation $F_{i}^t$ for any given time slice is obtained by concatenating one-hot representations of all the above features into single vector form $[G_i, R_i, A_i, h_i^t]$ where $h_i^t$ represents homogeneous feature representation for patient $i$ in time slice $t$, we denote its length as $n$. The input $F_i^t$ is passed through an embedding layer, where an embedding is learnt for each of feature, represented by $E$ of $n$$\times$$e$, where $e$ is embedding dimension and is then fed into the self-attention layer. We compute attention for every feature with respect to other features in the $F_i^t$ via $E$. The self-attention layer produces a vector of weights $a$: where $w_{s_1}$ is a weight matrix with a shape of $d_a$$\times$$e$ and $w_{s_2}$ is a vector of parameters with size $d_a$.
%Since $E$ is of dimension $n$$\times$$e$, the annotation vector $a$ will be of size $n$, and the softmax function ensures all the computed weights sum to 1.
To capture nuanced interactions especially for $F_i^t$'s with long sequences of $h_i^t$'s, we perform multiple hops of attention. As an example, say we want $r$ different parts to be extracted from the $F_{H_i}$, we extend $w_{s_2}$ into a $r$$\times$$d_a$ matrix, note it as $W_{s_2}$, and the resulting annotation vector $a$ becomes annotation matrix A. It's formally represented in equation \ref{eq:a_softmax}
where the softmax function is applied along the second dimension of its input. We compute the $r$ weighted sums by multiplying the annotation matrix $A$ and embedding output $E$ matrix resulting in $Q_i^t$ $=AE$.
\begin{equation}
    a = softmax ( w_{s_2} tanh ( w_{s_1} E^T)) \quad   \mbox{ and } \quad  A = softmax ( W_{s_2} tanh ( W_{s_1} E^T)) \label{eq:a_softmax}
\end{equation}
To capture the longitudinal dependencies and understand the importance of each time slice for a given patient, we feed the sequence of encoded quarterly representations from the self-attention layer into a bidirectional GRU-based RNN with aggregated MLP-Attention, refer to Appendix \ref{sec:mlp-attention} for details.

\textbf{Baselines:} We use ten baselines categorized in to common baselines (Logistic Regression, Random Forest, Multi-Layer perceptron), Deep Learning Baselines(1D-CNN and Bi-directional GRU based) and attention based models. A wide variety of baselines were evaluated in order to understand the performance vs model complexity trade off. Baselines are described in depth in Appendix \ref{attn_baseline}.

\section{Experiments}
We perform a robust evaluation with 11 onset prediction models on four clinical conditions (CHF, Kidney Failure, Diabetes Type II and COPD) over three axes (Performance, Generalization and Interpretability). Given the imbalance in data, we consider the Area under the Precision-Recall Curve (AUC-PR) as the primary metric for performance [\cite{saito2015precision}, \cite{davis2006relationship}] and is reported in Table \ref{tab:pr_results}. Standard deviations from three-fold cross validation is reported in Appendix \ref{app:perf}.

\textbf{\emph {Experiment i) Across clinical conditions}}: For each of four conditions mentioned above, we created a training set, validation, internal test (P1) and external test (P2) and use AUC-PR as the primary metric for evaluating performance.

\textbf{\emph {Experiment ii) Generalize across populations}}: Hospital systems can have variations in how diagnosis codes are assigned for each clinical visits as shown in many studies [\cite{burns2011systematic}, \cite{quach2010administrative}, \cite{jolley2015validity}, \cite{vlasschaert2011validity}], hence its essential for the model to be evaluated on different populations \cite{justice1999assessing}. Several studies show that characterizing performance on a single population can be insufficient [\cite{collins2014external}, \cite{bleeker2003external}, \cite{konig2007practical}, \cite{ivanescu2016importance}]. Hence to evaluate, we pick the same trained model evaluated on test set (P1) and evaluate it on corresponding condition's cohort in external cohort (P2) and report AUC-PR in Table \ref{tab:pr_results} under section P2.

\textbf{\emph {Experiment iii) Interpretablity}}: Non-linearity in deep learning based models help achieve better performance over linear methods, but may make model opaque to humans. In order to trust the model’s prediction, we believe alignment they should provide insights into why the model produced the result it did. We evaluate the interpretability of the models by generating both population level (Appendix \ref{app:pop_heatmaps}) and per patient feature importance visualizations for SAVEHR (Figure \ref{fig:heatmap}).

\begin{table}[htbp]
\centering
  %\caption{AUC-PR for all conditions in  population P1 and P2 with 15-month prediction window}
    \resizebox{0.85\textwidth}{!}{\begin{minipage}{\textwidth}
     \caption{Area under the curve (AUC-PR) performance across populations and conditions with 15-month prediction window}
   
    \begin{tabular}{lllllllll}
    \toprule
     &   P1 & & & &  P2  \\
    \midrule
    \textbf{AUC-PR} & CHF & KF & Diabetes & COPD & CHF & KF & Diabetes & COPD  \\   \midrule
    LR & 0.2636 & 0.2622 & 0.2560 & 0.2465 & 0.5708 & 0.2425 & 0.3182 & 0.1958 \\
    RF & 0.2729 & 0.3983 & 0.3154 & 0.6088 & 0.5211 & 0.2544 & 0.2421 & 0.1984 \\
    MLP & 0.4536 & 0.4771 & 0.4637 & 0.5491 & 0.5361 & 0.2204 & 0.4127 & 0.4588 \\
    BG & 0.4921 & 0.5969 & 0.4859 & 0.7594 & 0.712 & 0.279 & 0.5462 & 0.3731 \\
    CNN-1G & 0.5122 & 0.5565 & 0.4284 & 0.7371 & 0.6853 & 0.1741 & 0.3509 & 0.4204 \\
    CNN-LK & 0.5333 & 0.5809 & 0.4954 & 0.7413 & 0.6706 & 0.2839 & 0.4599 & 0.4503 \\
    BG-A & 0.4978 & 0.6009 & 0.5125 & 0.7436 & 0.6725 & 0.3395 & 0.6502 & 0.3662 \\
    Dense-A & 0.5109 & 0.5581 & 0.4745 & 0.7264 & 0.7099 & 0.3559 & 0.6523 & 0.385 \\
    CNN-1G-A & 0.5043 & 0.5330 & 0.4976 & 0.7380 & 0.6988 & 0.3959 & 0.5843 & 0.4405 \\
    CNN-LK-A & 0.5353 & 0.5464 & \textbf{0.5474} & 0.7734 & 0.7016 & 0.3743 & 0.6251 & 0.3562 \\
    SAVEHR & \textbf{0.5464} & \textbf{0.6112} & 0.5174 & \textbf{0.7776} & \textbf{0.7541} & \textbf{0.5819} & \textbf{0.7074} & \textbf{0.4839}
         \label{tab:pr_results}
    \end{tabular}
     \end{minipage}}
\end{table}

\section{Results and Discussion} 
The SAVEHR model outperformed all baselines models on AUC-PR metric across all four conditions on the internal test set P1 (except Diabetes in P1 and the external test set P2 as well. In the external test set P2, SAVEHR gains ranged from 7-46\% over the next best performing model as shown in \ref{tab:pr_results}. 

\textbf{External Test Cohort:}  A major strength of our work is that we used a formal external test cohort which is as large as most studies' development cohorts to validate the model's performance. Importantly, performance, as measured by AUC-PR, was higher (except Diabetes) in the external test cohort providing evidence that our architecture may generate models that generalize across cohorts. Although testing model performance seems to be an important criteria, a vast majority of published studies do not evaluate how their model transfers to a new population. 

\begin{wrapfigure}{L}{0.6\textwidth}

%\begin{figure}[htbp]
\centering
    \includegraphics[width=0.6\columnwidth]{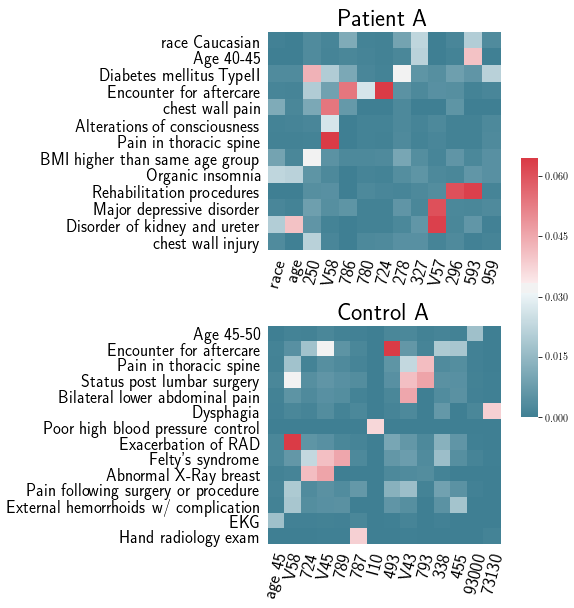}
     \caption{Feature importance heatmaps}
\label{fig:heatmap}
%\end{figure}

\end{wrapfigure}

\textbf{Interpretability:}  Predictive models are not, in general, intended to be explanatory, yet clinicians certainly desire an explanation of the model’s prediction particularly when that prediction is inconsistent with the clinician’s intuition. A powerful characteristic of the SAVEHR architecture is that it allows us to assign importance (or risk) scores to features and combination of features (Figure \ref{fig:heatmap}). While not a full explanation, we believe based on the findings in this study that it may be possible to provide the clinician with a summary and visualization that provides an indication of the underlying reasoning for model's prediction for an individual patient. In addition, by exploring the importance scores across populations of patients such as those in a certain age category or with specific risk predictions the clinician may gain insight into which features contribute to risk in that category of patients.

We examine the feature importance for two patients one with elevated Congestive Heart Failure (CHF) risk, Patient A (57\%.) and Control A (13\%) who correspondingly have similar characteristics (demographics and clinical encounters). We graphically illustrate the importance of pairwise feature interactions with color from deep blue to deep red indicating increasing importance. The features listed on the x-axis and the y-axis are the same for each panel, x-axis represents the ICD-9/10 code, while y-axis has descriptive labels for the codes. The mutual interactions are averaged, given there is no precedence for a feature over other. We observe that the patient identified as high risk has more interactions with high importance than the patients identified as low-risk, the interactions with high importance are multiple and diffuse: There are not one or two interactions but many that have high importance in patients with elevated risk, and many of the high importance interactions, but certainly not all, make clinical sense. A similar visualization is provided for one of the best performing attention based baselines in Appendix (Figure \ref{fig:cnn_hmap}).

\section{Conclusion}

We provide a new self-attention based deep neural network architecture to extract interpretable and actionable information from heterogeneous, sparse time-series data from electronic health records. We provide a multitude of performance metrics on the models for a comprehensive comparison of the current state-of-the-art and our models. Our model yields SoTA results across four different clinical conditions on an external cohort of thousands of patients monitored for a year or longer. Finally, we provide samples from anonymized patients to identify the interpretability of prediction scores to demonstrate how clinicians can incorporate our risk scores into the clinical workflows. We believe the relative importance of these features and feature interactions with a appropriate visualization can improve clinician's confidence in model predictions. Clinicians could utilize these predictions to target and modulate clinical interventions with greater precision.
\bibliography{ml4h_refs}  % .bib
\bibliographystyle{IEEEtran}

%## FOR AUC-ROC vs AUC PR

\newpage 
\appendix

%1. Population stats
%2. Disease breakdown
%3. attention architecture
%4. baselines
%5. Results
%6. Heatmaps
%7. feature importance

%%%% 1. Population Stats
\section{Population Statistics in P1 and P2}
\begin{figure}[htbp]
\centering
  \begin{subfigure}[b]{0.4\textwidth}
    \includegraphics[width=\textwidth]{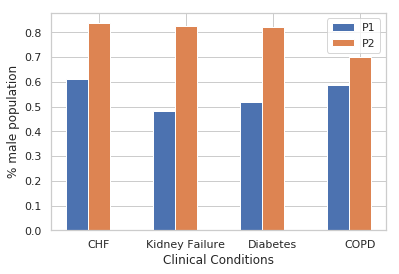}
    %\caption{Picture 1}
    %\label{fig:popstats1}
  \end{subfigure}
  \begin{subfigure}[b]{0.4\textwidth}
    \includegraphics[width=\textwidth]{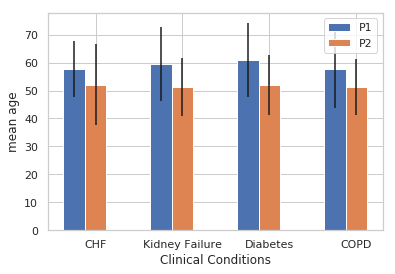}
    %\caption{Picture 2}
    %\label{fig:popstats2}
  \end{subfigure}
  \caption{Population statistics for P1 and P2}
  \label{fig:popstats}
\end{figure}

We describe the population statistics such as gender ratio and average age for all the four clinical conditions and present them in Figure \ref{fig:popstats}.

%%% COHORTS
\section{Disease Cohorts}

We create cohorts (Training, Validation and Test) on Population P1, and use P2 entirely as an external test set for four chronic diseases - Congestive Heart Failure (CHF), Kidney Failure, Diabetes Type II and Chronic Obstructive Pulmonary Disease (COPD).

\label{index_date}
\begin{table}[htbp]
 \centering 
 \resizebox{0.95\textwidth}{!}{\begin{minipage}{\textwidth}
  \caption{Disease cohorts with train, validation and test sets for P1 and external test (P2) populations} 
 \centering
\begin{tabular}{@{}lllll@{}}
\toprule
Disease  & Training  (P1) & Validation (P1) & Test (P1) & External Test (P2) \\
            &  &         case : control & & \\ \midrule 
CHF & 14343 : 159567 & 793 : 8361 & 3916 : 41851 & 1259 : 5890 \\
Kidney & 8085 : 66045 & 447 : 3455 & 2216 : 17292 & 757 : 9351 \\
Diabetes Type II & 7674 : 53308 & 429 : 2781 & 2088 : 13961 & 3422 : 6997 \\
COPD & 11301 : 104719 & 641 : 5466 & 3107 : 27425 & 1767 : 5000 
\label{tab:popstats}
\end{tabular}
    \end{minipage}}
\end{table}

%%% 2. Disease Cohorts
\section{Index Date}

\begin{figure}[h]
\centering
    \includegraphics[width=0.8\columnwidth]{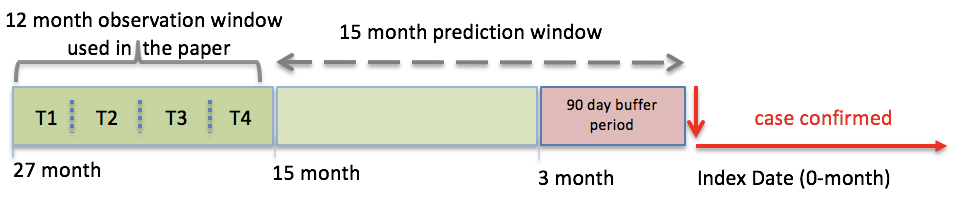}
     \caption{Illustration of index date, observation and prediction windows}
     \label{fig:index_date} 
\end{figure}

The case-control design within cohorts for each disease was created for patients who received care between 2015 and 2018. Incident cases for each condition was defined as patients between ages of 30 and 80 years of age for whom an ICD-9 or ICD-10 code representing the condition was recorded as an encounter diagnosis at least three times in a six-month period but never had any prior diagnosis for the condition. We defined the index date as the date of the first of the three qualifying encounters. We did not consider any of the data from the 3 months prior (buffer period) to the index date in order to avoid incorporating diagnostic data that had been obtained but not yet resulted in a diagnosis being recorded. Control patients were selected as those who had at least 5 encounters in a 2-year period but never had a diagnostic code for the condition being modelled recorded. The last encounter recorded in the system was chosen as index date for control patients. Figure \ref{fig:index_date} illustrates our use of 12-month observation window that's between 27 and 15-months from the index date and a prediction window of 15-months. Since, the observation window was fixed across a relatively long time window (12-month), we aggregate frequency counts of diagnosis codes assigned for encounters across a time window similar to that in \cite{Choi2016UsingRN} to facilitate temporal learning. Codes that had fewer than 50 occurrences in cohort were filtered out. 

\section{End-to-End Data and Modeling Pipeline}
HealtheDataLab is a big data processing platform built on Amazon EMR. The data, population health data with longitudinal patient records are ingested from Amazon S3. The end-to-end flow is as follows, an AWS Data Pipeline job orchestrates the transformation of data, the launch of an Amazon EMR cluster, and creates a data catalog along with a Hive metastore in Amazon RDS and AWS Glue. HealtheDataLab provides a Jupyter notebook running on an EC2 instance that connects to a spark pipeline on Amazon EMR. HealtheDataLab has custom packages like ontologies, FHIR support, and concepts mapping to empower data scientists to create patient cohorts in a very simplified manner. Once cohorts are created, they are stored in S3 as compressed numpy arrays. Then, the Amazon SageMaker machine learning job is kicked off with specified cohort location in S3, along with hyper-parameters to be optimized for. After the completion of the job, the best hyper-parameters are recorded and the job id is noted to run evaluation on secondary populations.    

\begin{figure}[htbp]
\begin{center}
    \includegraphics[width=0.75\textwidth]{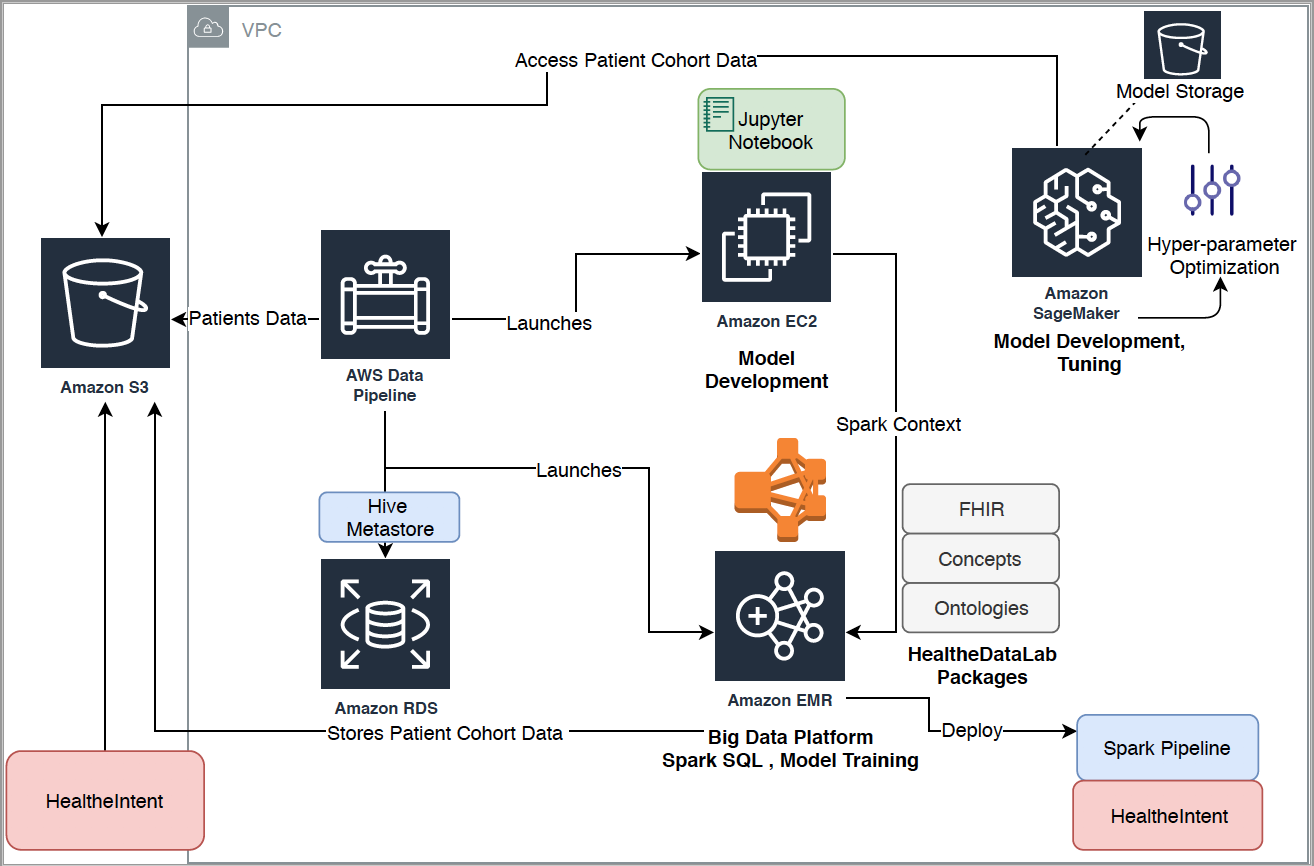}
   \caption{HealtheDataLab Workflow on AWS}
\end{center}
\end{figure}

%% 3. Attention
\section{Aggregated Deep Representation with MLP Attention across quarters}
\label{sec:mlp-attention}
To capture the longitudinal dependencies and understand the importance of each quarter for a given patient, we feed the sequence of encoded quarterly representations from the self-attention layer into a bidirectional GRU-based RNN, presented in Equation \ref{eq:gruattn}
\begin{equation}
    h_1, h_2, h_3, h_4 = BiGRU(Q_i^1, Q_i^2, Q_i^3, Q_i^4) \label{eq:gruattn}
\end{equation}
where $h_t \in R $ represents the output by the GRU for quarter $t$. We use MLP Attention (or multi-layer perceptron attention) [58] on top of the BiGRU layer to obtain weighted representation of each quarter.The weight of each attention vector $t$ at GRU output $t$, $\alpha_{ti}$, is calculated as a normalized weighted sum,
\begin{equation}
    %\alpha_{ti} = \frac{\exp$e_{t_i}$)}{ \Sigma_{k}^L \exp($e_{t_k}$) }   \label{eq:alphati}
    \alpha_{ti} = {e^{e_{t_i}}}/{\Sigma_{0}^4 e^{e_{t_k}}} \mbox{ where } e_{ti}  =  f_{att} (\mathbf{a}_i, \mathbf{h}_{t}), \label{eq:eti}
\end{equation}
where $\mathbf{h}_{t}$ are hidden state vectors from the BiGRU cell, $\mathbf{a}_i$ the attention network and $f_{att}$ an attention model. Once we obtain the attention weights, the vector representation aggregated $V_{F_i}$ for the $i^{th}$ patient across quarters is computed by:
$$V_{F_i} = \Sigma_0^4 \alpha_{t_i}* h_{t_i}$$
Once we obtain the aggregated representation for a patient across quarters, we add a softmax layer for the final outcome prediction given by 

$ \hat{y} = softmax( W_{savehr}^T * V_{F_i} + b_{savehr}) $

%% 5. Baselines
%%% BASELINES

\begin{figure}[htbp]
\section{Baseline Models}
   \label{attn_baseline}
%\begin{wrapfigure}{L}{0.5\textwidth}
\begin{center}
    \includegraphics[width=0.75\textwidth]{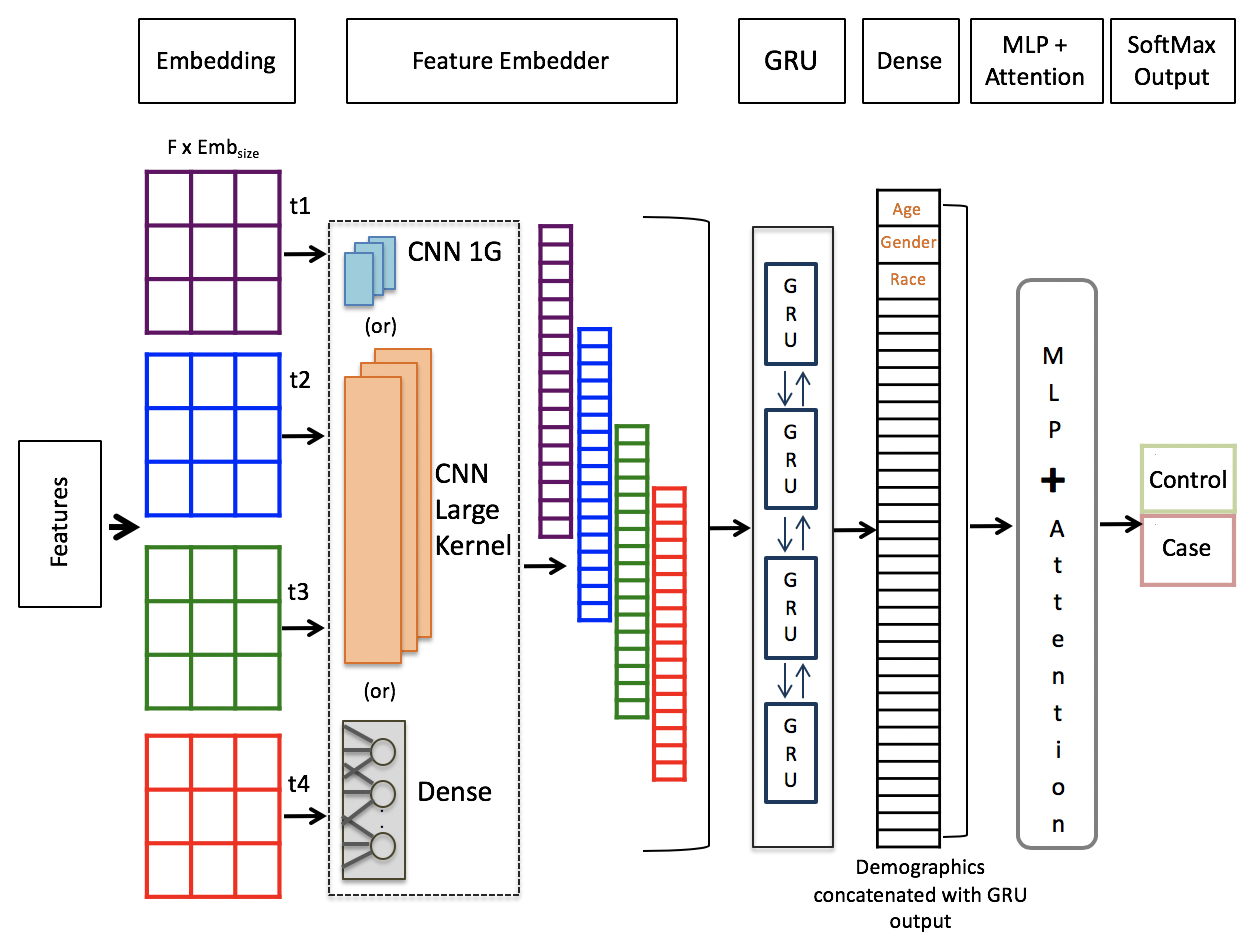}
   \caption{Attention based Architecture}

\end{center}
\end{figure}
%\end{wrapfigure}

\textbf{Logistic regression and Random Forest:}  We trained three commonly used baselines - logistic regression (LR), random forest (RF) and a multi-layer perceptron (MLP) with dropout. For the logistic regression and random forest, we use the implementation provided by scikit-learn\cite{scikit-learn}with no regularization. 

\textbf{Deep Learning Baselines} \emph{(1D-CNN + BiGRU and BiGRU)}\textbf{:} Inspired by the success of 1D-CNN and RNN based architectures for clinical notes \cite{liu2018deep}, we extend that to structured EHR data. The diagnosis codes for conditions and procedures that we collectively term as medical concepts, can be considered analogous to words in sentences. We use an embedding layer to encode the features into a continuous space. We use frequency of each code assigned in a time slice, and concatenate the frequencies for each code into the medical concept embedding as shown to be effective in \cite{mallya2019effectiveness}. We feed the embedding data into a 1D-CNN first. Since the medical concepts are inherently not ordered, we use the equivalent of 1-gram, i.e a kernel size of 1 across feature embeddings for the 1D-CNN on each of the time slices. To exploit the longitudinal nature of EHR data, we feed the time slice aggregated representation from 1D-CNN into a bidirectional GRU (Bi-GRU) layer, we name this model CNN-1G. To measure the incremental value of 1D-CNN filters we create another baseline (BG) that uses only a Bi-GRU layer on top of the embedding layer. Demographic information may be static in nature, but is very critical to clinical decisions, hence we incorporate these features by concatenating to the Bi-GRU layer output from both the models described earlier.We also experiment with  BiGRU instead of 1D-CNN and report performance on that. 

\textbf{Attention based Deep Learning Baselines} \emph{(CNN-LargeKernel + BiGRU + Attention)}\textbf{:} To understand if attention could help in EHR based modeling, we add MLP attention \cite{chorowski2015attention} to the BiGRU layer for the baselines (CNN-1G \& BG) described in the earlier section to enable them to focus on the most important time window. The approach above with 1D-CNN wouldn't capture the interactions among features very effectively due to our kernel size of 1.Hence, ideally we'd like to compute for any 2,3 or n-grams of features their collective and relative importance. Given that the ordering of medical concepts within a given time window doesn't matter, anything beyond a 1-gram kernel would require an ordering. To incorporate this and avoid the need for massive n-gram computation, we propose a novel baseline named CNN Large Kernel (CNN-LK), where the 1D-CNN kernel size is set equal to the number of input features, essentially giving us a weighted combination of all the input features. To understand if the 1D-CNN adds values, we also use another baseline where we replace the large kernel layer with a Dense layer of the same size. We note that, for the aforementioned architectures, we are unable to determine pairwise importance between any two features. 

%% 5. Results AUC-ROC
\section{AUC-ROC results for all conditions across populations}

\begin{table}[htbp]
    \centering

   \resizebox{0.85\textwidth}{!}{\begin{minipage}{\textwidth}
    
    \caption{Area under the curve (AUC-ROC) performance across populations and conditions with 15-month prediction window}

    \begin{tabular}{lllllllll}
    \toprule
     &   P1 & & & &  P2  \\ \midrule
   \textbf{AUC-ROC} & CHF & KF & Diabetes & COPD & CHF & KF & Diabetes & COPD  \\   \midrule
    LR & 0.8474 & 0.8586 & 0.8358 & 0.8628 & 0.7949 & 0.7441 & 0.7426 & 0.8187 \\
    RF & 0.8187 & 0.8314 & 0.7980 & 0.8826 & 0.8138 & 0.7159 & 0.7038 & 0.6466 \\
    MLP & 0.8466 & 0.7969 & 0.8170 & 0.8435 & 0.8167 & 0.8271 & 0.7861 & 0.8341 \\
    BG & 0.8695 & 0.8677 & 0.8411 & 0.9129 & 0.8497 & 0.8066 & 0.7611 & 0.8325 \\
    CNN-1G & 0.8677 & 0.8684 & 0.8247 & 0.9008 & 0.8724 & 0.8407 & 0.8611 & 0.8137 \\
    CNN-LK & 0.8717 & 0.8709 & 0.8641 & 0.9144 & 0.8661 & 0.7983 & 0.8505 & 0.8437 \\
    BG-A & 0.8751 & 0.8725 & 0.8467 & 0.9123 & 0.8724 & 0.8382 & \textbf{0.8922} & 0.8543 \\
    Dense-A & 0.8695 & 0.8628 & 0.8392 & 0.9084 & 0.8769 & 0.8273 & 0.8726 & 0.7934 \\
    CNN-1G-A & 0.8722 & 0.8575 & 0.8409 & 0.9055 & 0.8909 & 0.8424 & 0.8651 & 0.8362 \\
    CNN-LK-A & \textbf{0.8752} & 0.8609 & 0.8421 & 0.9068 & 0.8811 & 0.8584 & 0.8913 & 0.8081 \\
    SAVEHR & 0.8749 & \textbf{0.8728} & \textbf{0.8717} & \textbf{0.9160} & \textbf{0.9093} & \textbf{0.8616} & 0.8788 & \textbf{0.8369}
    \end{tabular}

\label{tab:roc_results}
\end{minipage}}
\end{table}

%Average Attention per quarter 
%[0.22868084, 0.2113695 , 0.20590293, 0.35404657]
\section{Distribution of MLP attention weights across quarters}
In order to understand the importance across time slices, we compute the average attention per time slice across entire test set P1 and report it below in \ref{tab:att_ts}. T4, the closest to the index date is the most prominent quarter.

\begin{table}[htbp]
\begin{tabular}{lllll}
\toprule
Average Attention per timeslice  & t1 & t2 & t3 & t4 \\ \midrule
& 0.22868084 & 0.2113695 &  0.20590293 & 0.35404657 \\
\bottomrule
\end{tabular}
\caption{attention over quarters}
\label{tab:att_ts}
\end{table}

\subsection{MLP attention weights vs number of diagnosis counts} 
\begin{figure}[h]
\centering
  \begin{subfigure}[b]{0.4\textwidth}
    \includegraphics[width=\textwidth]{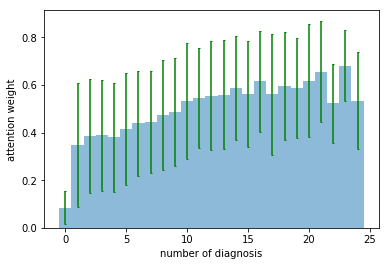}
    \caption{For predicted case patients}
    \label{fig:case_att}
  \end{subfigure}
  \begin{subfigure}[b]{0.4\textwidth}
    \includegraphics[width=\textwidth]{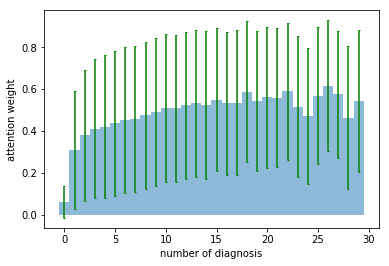}
    \caption{For predicted control patients}
    \label{fig:control_att}
  \end{subfigure}
  \caption{MLP attention scores vs number of diagnosis in a timeslice}
\end{figure}

To assess the importance of attention with respect to the number of diagnosis in a given time-slice, we plot the average and standard deviation for the diagnosis counts. We observe that the model very low to zero attention to quarters without any diagnosis code. As the count increases, attention increases but the large error bars in both Figure \ref{fig:case_att} and Figure \ref{fig:control_att} suggest that its not always paying attention to time-slice with the most counts.  

\section{Performance metric graphs with error bars}
\label{app:perf}
\begin{figure}[htbp]
\centering
    \includegraphics[width=\columnwidth]{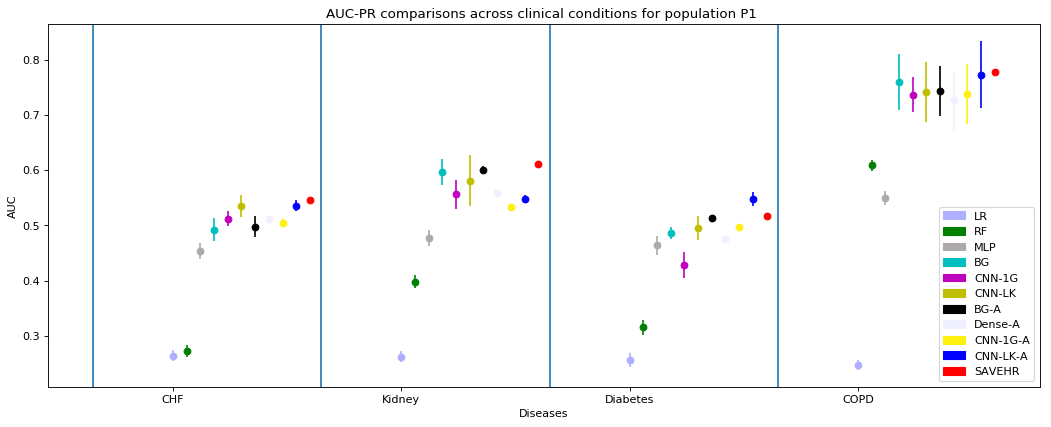}
     \caption{AUC-PR in P1 for all models with cross validation error bars}
\label{fig:apr}
\end{figure}

\begin{figure}[htbp]
\centering
    \includegraphics[width=\columnwidth]{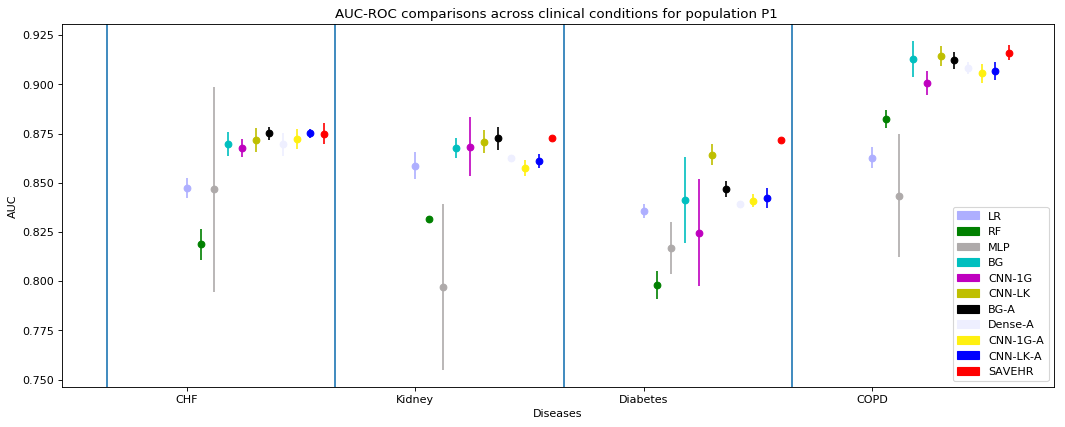}
     \caption{AUC-ROC in P1 for all models with cross validation error bars}
     \label{fig:aproc}
\end{figure}

In this section, we report the AUC-PR (Figure \ref{fig:apr}) and AUC-ROC (Figure \ref{fig:aproc}) for population P1, with cross validation error bars. We observe that SAVEHR, CNN-1G-A and CNN-LK-A have very low standard deviations for both AUC-PR and AUC-ROC across all diseases when compared to the other models.

\section{Example patient heatmaps for CNN-LK-A model}
In section 5 of the paper, we provide an example visualization for the SAVEHR model. To contrast that, below we provide heatmap from the CNN-LK-A model on the same set of case and control patients.

\label{app:lka_heatmaps}
\begin{figure}[h]
\centering
    \includegraphics[width=\columnwidth]{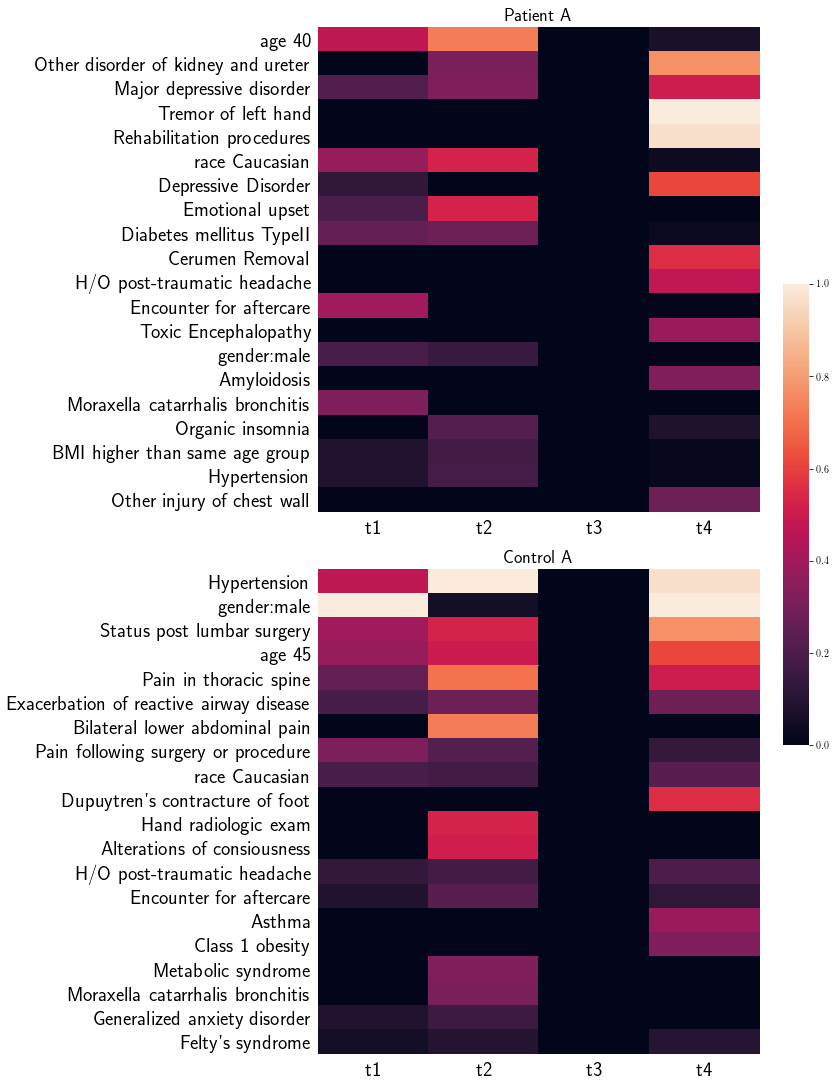}
     \caption{Feature importance for CNN-LK-A}
     \label{fig:cnn_hmap}
\end{figure}

\newpage 
\section{SAVEHR Case population heatmaps in P1 and P2 for CHF}
\label{app:pop_heatmaps}

\begin{figure}[htbp]
\centering
    \includegraphics[scale=0.28]{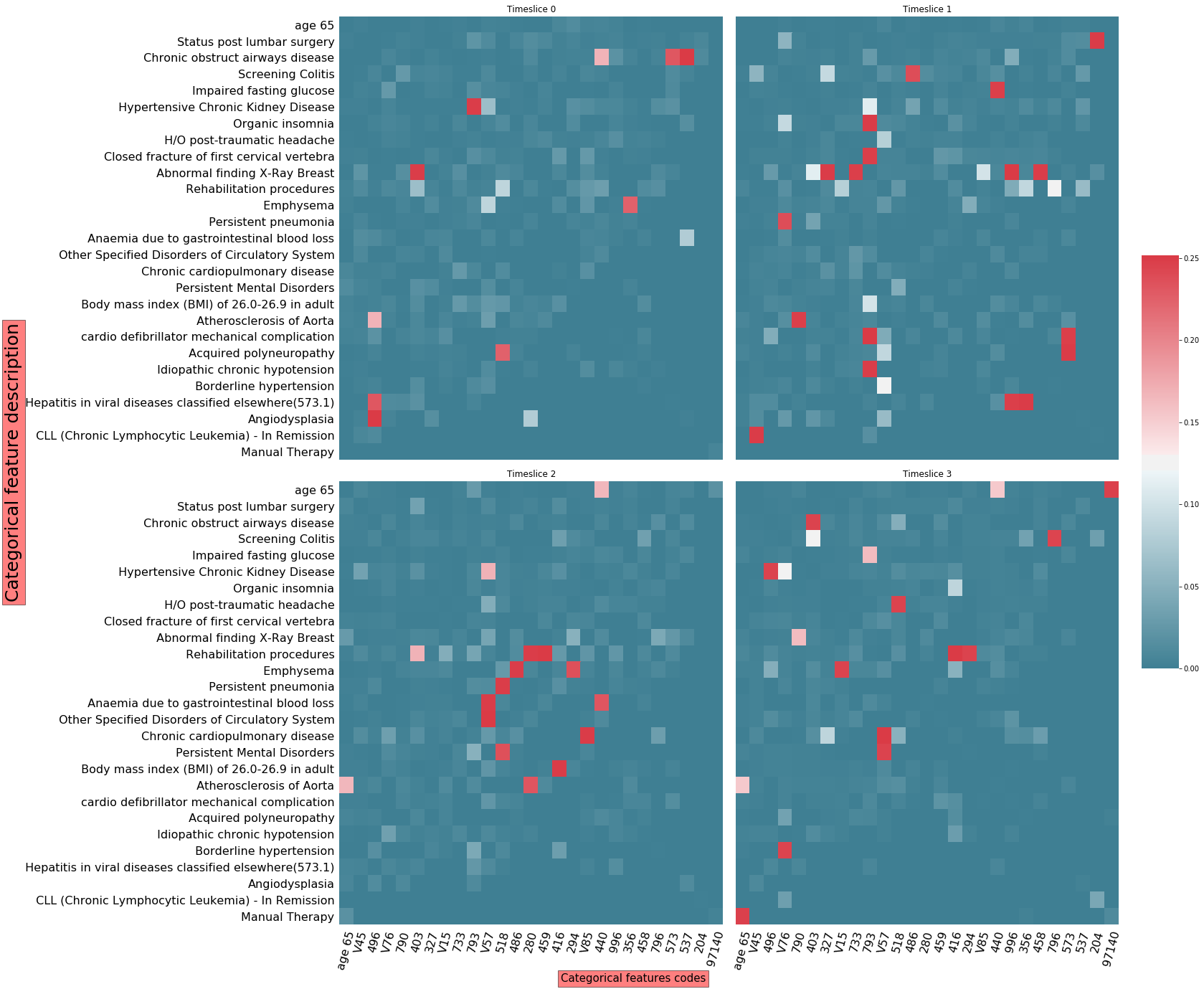}
     \caption{SAVEHR heatmap visualization across quarters for all case patients in P1}
\label{app:p1_heatmaps}

\end{figure}

\begin{figure}[htbp]
\centering
    \includegraphics[scale=0.28]{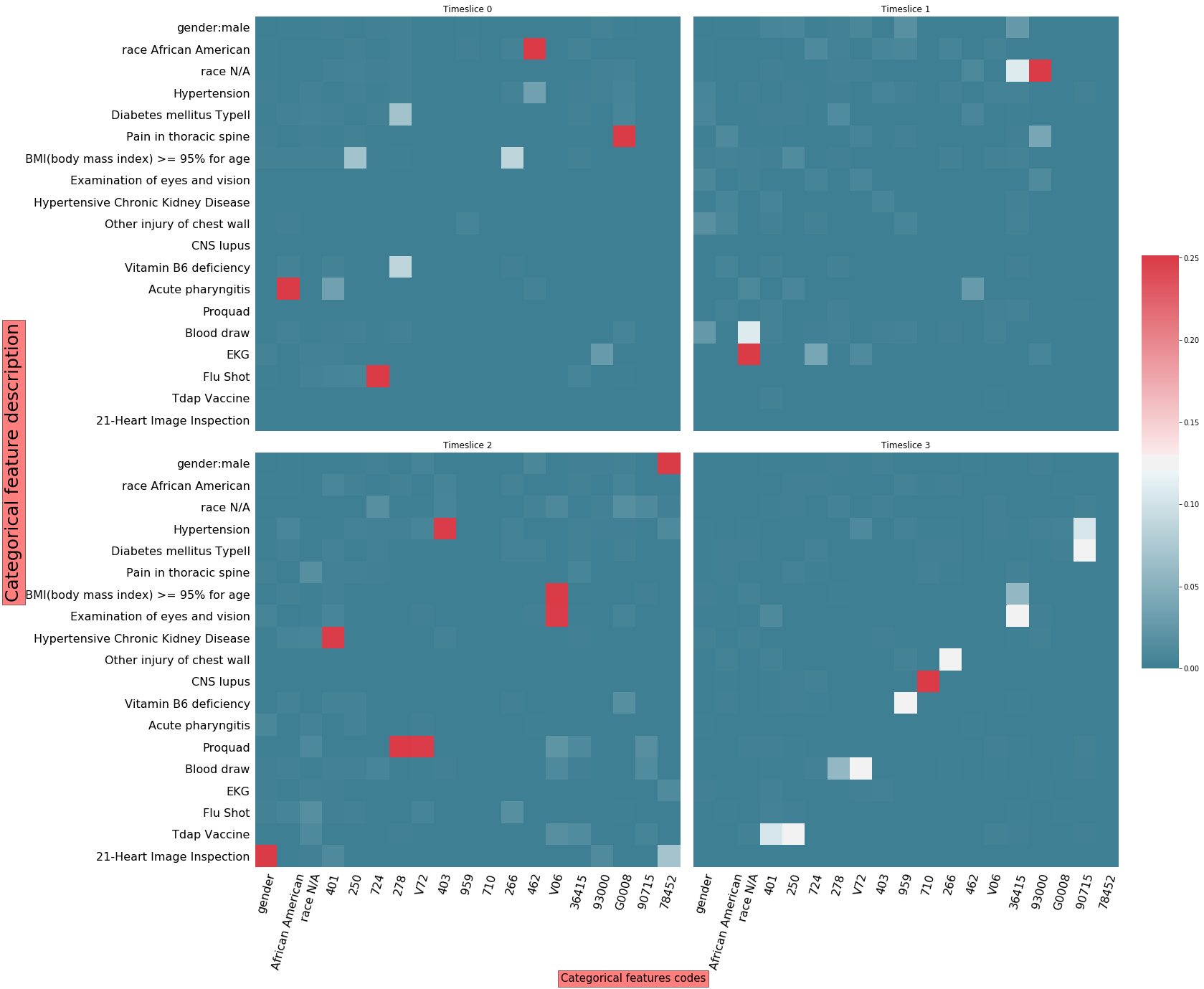}
     \caption{SAVEHR heatmap visualization across quarters for all case patients in P2}
\label{app:p2_heatmaps}

\end{figure}

To understand the features that induce risk across the population as a whole, we generate averaged heat maps across all the case patients in P1 (Figure \ref{app:p1_heatmaps}) and P2 (Figure \ref{app:p2_heatmaps}). Noticeably, the top features in both of the populations differ, suggesting that the model is able to learn different characteristics and adapt. 

\newpage
\section{Feature importance tables for baselines }
We report the feature importance as determined by averaging the importance scores predicted by the model across the predicted case patients.

\begin{table}[htbp]
\begin{tabular}{@{}llll@{}}
\toprule

Logistic Regression & LR coefficient & Diagnosis Code & Description\\ \midrule
 & 1.42 & 90656  & Flu Vaccine \\ 
 & 1.207 & G0378 & Hospital Observation Service\\
 & 1.096 & 735 & Acquired hammer toe\\
 & 0.934 & 361 &  Retinal defect\\
 & 0.926 & v54 &  Aftercare fracture arm\\
 & 0.901 & 816 &  Closed fracture of middle phalanx of second finger of right hand\\
 & 0.862 & 557 &  Enterocolitis\\
 & 0.794 & 191 & Malignant Neoplasm of Brain\\
 & 0.788 & 041 & Mycoplasma infection in conditions classified elsewhere \\
 &  0.783 & 432 &  Chronic spont intraparenchymal hemorrhage \\

 \bottomrule
\end{tabular}
 \end{table}

\begin{table}[htbp]
    \begin{tabular}{@{}llll@{}}
    \toprule
    Random Forest & coefficient & Diagnosis Code & Description\\ \midrule
    & 0.035 & age   & Age\\ 
    & 0.006 & race   & Race\\
    & 0.006 & v58   & Encounter for other and unspecified procedures and aftercare\\
    & 0.005 & 401   & Essential Hypertension \\
    & 0.005 & v76   & Screening Colitis\\
    & 0.005 & 36415   & Blood Draw\\
    & 0.004 & v57   & Care involving use of rehabilitation procedures\\
    & 0.004 & gender   & Gender \\
    & 0.004 & v70   & General psychiatric examination \\
    & 0.004 & 786   & Chest wall pain\\
    \bottomrule
\end{tabular}
 \end{table}
 
\newpage

\begin{table}[htbp]
    \begin{tabular}{@{}llll@{}}
    \toprule
    CNN 1 gram & Importance score  & Diagnosis Code & Description\\ \midrule
    & 0.060405 & v76 & Screening Colitis \\ 
    & 0.043208 & 427   & Atrial tachycardia\\
    & 0.041183 & v45   & Status post lumbar surgery\\
    & 0.03769 & 793   & Abnormal Findings X-Ray Breast \\
    & 0.02934 & v57   & Care involving use of rehabilitation procedures\\
    & 0.020804 & 585   & chronic renal failure\\
    & 0.018625 & 562   & Small bowel diverticular disease\\
    & 0.017839 & 530   & Cardiochalasia\\
    & 0.016736 & v10   & Personal History of Malignant Neoplasm of Eye\\
    & 0.016391 & 455   & External hemorrhoids with complication\\
     \bottomrule
    \end{tabular}
 \end{table}

\begin{table}[htbp]
\begin{tabular}{@{}llll@{}}
\toprule

CNN LargeKernel & Importance score  & Diagnosis Code & Description\\ \midrule
& 0.05082 & 569 & Colostomy and enterostomy complications \\ 
& 0.48861 & M06   & Rheumatoid arthritis with negative rheumatoid factor (HCC)\\
& 0.38591 & 333   & degenerative diseases of the basal ganglia \\
& 0.35872 & R57   & Cardiogenic Shock \\
& 0.34976 & C95   & Acute leukemia\\
& 0.31523 & 250   & Diabetes mellitus TypeII\\
& 0.31516 & I62   & Nontraumatic subdural hemorrhage\\
& 0.31463 & 182   & Malignant Neoplasm of body of uterus\\
& 0.29420 & H25   & Senile cataract of right eye\\
& 0.29093 & M21  & Limb deformity\\
 \bottomrule
\end{tabular}
 \end{table}

\end{document}